\documentclass{bmvc2k}

\usepackage{microtype}
\usepackage{graphicx}
\usepackage{subfigure}
\usepackage{booktabs} 

\usepackage{amsmath}
\usepackage{algorithmic}

\usepackage[dvipsnames]{xcolor}

\usepackage{tikz}
\usepackage{pgfplots}

\title{Bag of Negatives for Siamese Architectures}

\addauthor{Bojana Gaji\'c}{bgajic@cvc.uab.es}{123}
\addauthor{Ariel Amato}{aamato@vintra.io}{1}
\addauthor{Ramon Baldrich}{ramon@cvc.uab.es}{23}
\addauthor{Carlo Gatta}{cgatta@vintra.io}{1}

\addinstitution{
 Vintra Inc\\
 San Jose, California, USA
}
\addinstitution{
 Autonomous University of Barcelona\\
 Bellaterra,
 Barcelona, Spain
}
\addinstitution{
 Computer Vision Center\\
 Bellaterra,
 Barcelona, Spain
}

\runninghead{Gaji\'c et al}{Bag of Negatives for Siamese Architectures}

\begin{document}

\maketitle

\begin{abstract}
Training a Siamese architecture for re-identification with a large number of identities is a challenging task due to the difficulty of finding relevant negative samples efficiently. In this work we present Bag of Negatives (BoN), a method for accelerated and improved training of Siamese networks that scales well on datasets with a very large number of identities. BoN is an efficient and loss-independent method, able to select a bag of ``high quality negatives'', based on a novel online hashing strategy.
\end{abstract}

\section{Introduction}
Instance retrieval and re-identification (re-ID) are ranking problems where, given an image of an object, a system is searching for the images of the same object, usually taken by a different camera, under different light, from different angles etc. Several research lines address this problem \cite{guo2018efficient, facenet2015, zhao2017deeply, liu2017end, li2018harmonious, zhao2017spindle,su2017pose,saquib2018pose,xu2018attention}. Some of these works propose methods that are computationally very expensive, using network architectures that are task-specific \cite{zhao2017deeply, liu2017end, li2018harmonious, zhao2017spindle,su2017pose,saquib2018pose,xu2018attention}. Others focus on keeping architectures simple and flexible to adapt to any ranking problem \cite{facenet2015, 100Kids2017, hermans2017defense}.

Once the network architecture is designed, the question that naturally arises is: what should be the appropriate loss function for training? Several works use classification loss, due to its simplicity and the availability of several tools ready for the task \cite{zheng17unlabeled, Kalayeh_2018_CVPR}. These methods typically train the network as if every identity in the dataset is a class, while extracting image descriptors by pooling features from the last convolutional layer at testing time. Even though these approaches are very simple, they suffer from serious limitations in terms of the number of classes they can handle. For instance, a dataset with more than 10M identities (such as the FaceNet dataset, as reported in \cite{MegaFace2016}) and with, e.g, a $128$ embedding size, would require a fully connected layer with $1.28$ billion extra parameters, which would be impossible to train from a practical perspective.

Other approaches propose alternative functions, widely known as ranking losses \cite{taigman2014deepface, facenet2015, chen2017beyond}. The main goal of these losses is to bring the representations of samples from the same ID close to each other, while separating the representations of images that belong to different classes. To do so, Siamese architectures with several parallel streams, made of the same networks with shared weights, are used \cite{facenet2015, 100Kids2017, hermans2017defense}. These architectures allow computing compact representations for several input images simultaneously, combining them with a ranking loss. Even though these approaches do not depend on the number of classes in the training set, they are extremely hard to train due to the complexity of finding input pairs, triplets or quadruplets that produce a loss greater than zero. In this paper we propose an online strategy for mining samples, which contributes to a more efficient training of Siamese architectures.
\subsection{Relevant work}
Online hard negative mining (OLHN) methods take advantage of the sample representations available at mini-batch level in order to improve the probability of retrieving relevant negatives for the triplet loss \cite{oh2016deep,facenet2015,hermans2017defense}. The latter two methods will be discussed more in detail in the next section since they are the most popular (\cite{ge2018deep, samplingmatters2017iccv, 100Kids2017}) and most effective OLHNs, and we aim at improving their performance.

An offline re-weighting of the loss can improve the quality of negative samples, but at non-negligible computational cost \cite{samplingmatters2017iccv}. Taking advantage of extra knowledge on sub-categories within the dataset is also advantageous in mining negative samples \cite{Fine_Grained_2014}.

In recent years, online methods that provide relevant negative samples prior to triplet formation have been proposed, thus not at mini-batch level.

In \cite{ge2018deep} the authors propose a strategy that builds a tree of identities to facilitate the sampling of relevant negatives for a given anchor. The method clearly improve the quality of the negative samples but at the cost of updating the tree at every epoch. Also, the tree construction is based on an identity-to-identity distance matrix, which thus scales polynomially with the number of identities.

In \cite{100Kids2017} the authors explicitly face the problem of training a Siamese network with 100k identities. The basic idea is to generate a representation for each identity, and apply clustering on all the identities to generate clusters or subspaces, wherein identities are similar in each subspace. The authors propose to train a classifier on a subset of identities, then use the classifier to generate image representations, and finally perform k-means clustering on them in order to form the subspaces. The authors do not update the cluster during the training, thus the subspaces could become sub-optimal in later stages of the training.
\section{Motivation}

The triplet loss (see Equation (\ref{eq:triplet})) is based on the construction of triplets $i \in \mathcal{T}$ formed by an anchor sample $x_i^a$, a positive sample $x_i^p$ (belonging to the same ID as the anchor) and a negative sample $x_i^n$ (belonging to a different ID). The samples are mapped into an embedding by a given function $f(\cdot)$, that is usually a deep convolutional network, whose parameters are learned by minimizing the loss $\mathcal{L}$. The goal of the triplet loss is to ensure that the anchor-negative pairs are far from each other by a margin $\alpha$ with respect to the anchor-positive pair distance. The most challenging part of using the triplet loss is finding triplets that produce a non-zero loss. This is hard, since the number of all possible triplets in the dataset is proportional to the cube of total number of images $N$, $|\mathcal{T}| \sim N^3$, and the more the system trains, the less probable it is to find a relevant negative for a given anchor-positive pair.
\begin{equation}
   \label{eq:triplet}
   \mathcal{L} = \sum_{i \in \mathcal{T}} \left [ ||f(x_i^a) - f(x_i^p) ||_2^2 -  ||f(x_i^a) - f(x_i^n) ||_2^2 + \alpha \right ]_+
\end{equation}
Let $n$ be the average number of images per ID, $m$ the mini-batch size, $b$ number of triplets per mini-batch ($b = m/3$), and $k$ the number of images of each ID in the mini-batch. We introduce the notation $\hat{n}$, the number of negative samples that produce a non-zero loss if used in conjunction with the triplet loss and an anchor-positive pair. The ``quality'' of the retrieved negatives is also relevant, as pointed out in \cite{samplingmatters2017iccv}: negative samples have to be selected such that the anchor-negative distance distribution is almost uniform. 

Random negative sampling from the whole dataset (to which we refer as Vanilla sampling) has complexity $\mathcal{O}(1)$ but does not provide relevant negative samples except in the beginning of the training, since the probability of picking a relevant negative, $p_{\hat{n}} = \hat{n} / (N - n) \simeq \hat{n} / N$. From now on we will omit $n$ from the formula since it is negligible w.r.t. $N$.

\emph{Semi hard} loss \cite{facenet2015} employs a negative sampling strategy that has an increased cost due to the fact that the additional computed distances scale polynomially with the mini-batch size. The improvement in $p_{\hat{n}}$ with respect to Vanilla sampling is linearly dependent on the number of triplets $b$. For this reason, authors use huge mini-batches in the order of 1800 samples. $p_{\hat{n}}$ is thus increased at the cost of large mini-batches and additional sample to sample distance computation.

\emph{Batch hard} loss \cite{hermans2017defense} is an improved version of the \emph{semi hard} loss where, thanks to a more controlled mini-batch creation and additional distances computation, the method exhibits $p_{\hat{n}} = 3b \hat{n} / N$. This strategy offers a $50\%$ improvement in $p_{\hat{n}}$ w.r.t. the \emph{semi hard} approach, but still provides a probability that depends on the mini-batch size.

Finally, an offline exhaustive search into the dataset provides $p_{\hat{n}} = 1$. This is, of course, not viable for large datasets. Nonetheless, for relatively small datasets and with the proper sampling strategy over the $m N$ distances, exhaustive search provides excellent negatives samples \cite{samplingmatters2017iccv}.

\section{Contribution}

The paper's \textit{main contribution} is a \textbf{loss-independent} and \textbf{computationally inexpensive} \textbf{online} strategy for \textbf{improved negative mining} in large datasets, which we named Bag of Negatives (BoN). The main advantages of BoN w.r.t. previous methods are: (1) faster training due to more non-zero loss triplets; (2) better performance on validation sets due to high quality negatives; and (3) a negligible additional computational cost w.r.t. the Siamese architecture training.

BoN does not require computing additional sample representations nor their respective distances to be able to select an appropriate negative. Also, it does not require additional networks and/or losses to be applied on the input image, nor ad-hoc strategies. Moreover, it can be combined with any loss that requires a negative sample. Nonetheless, for simplicity, we will show the behaviour of BoN with triplets; analogous results are achieved with other ranking losses, such as, e.g, contrastive loss. We also want to emphasize that this approach is devised with large datasets and computational efficiency in mind. Finally, the method has only one relevant meta-parameter, discussed in \ref{sect:experiments}.
\section{Bag of Negatives}

\begin{figure}
\begin{center}
  \includegraphics[width=0.8\linewidth]{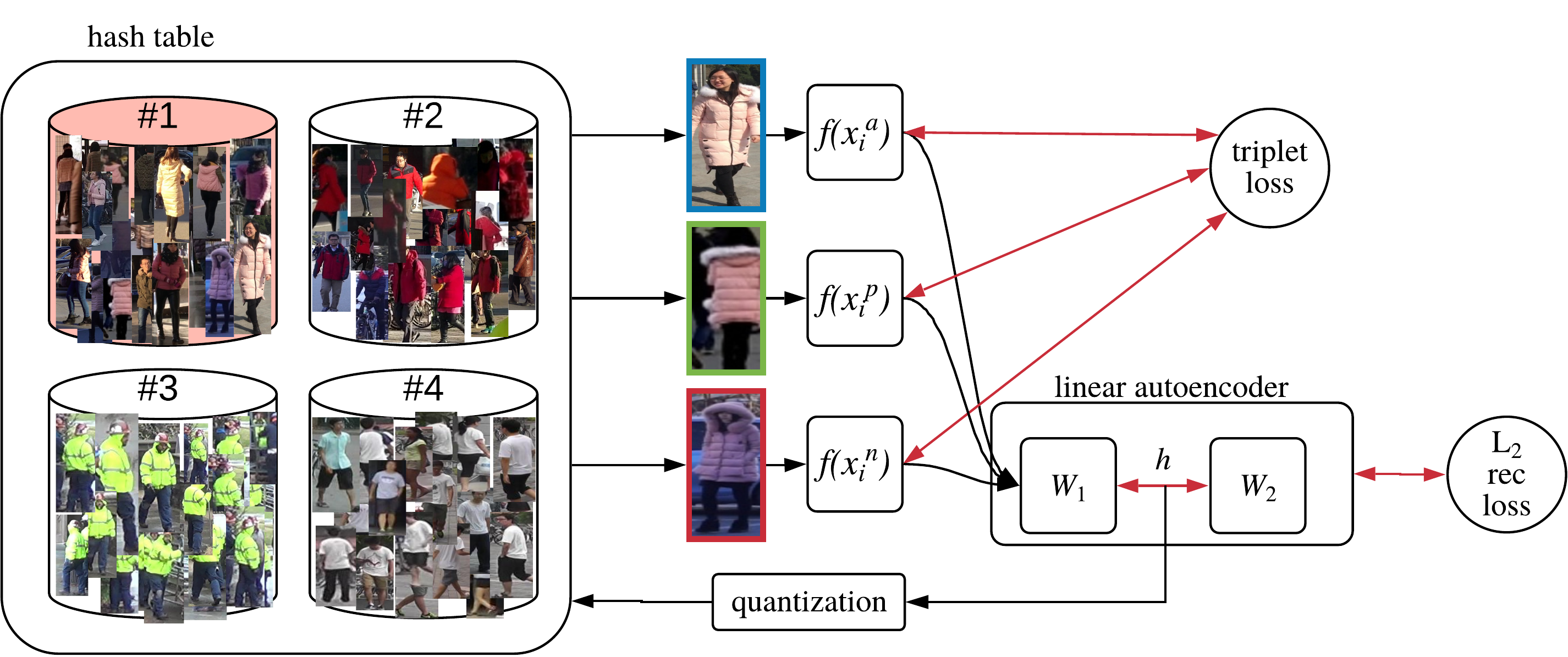}
  \caption{BoN strategy. Triplets with good quality negatives are formed using the information from the hash table. The resulting embedding is used to learn both the deep model $f$ and a linear projection that, in turn, provides a low-dimensional embedding. Its quantization provides (possibly) new entry positions in the hash table for the input images. The hash table and the linear autoencoder are updated at each training step with minimal overhead.}
  \label{fig:main}
\end{center}
\end{figure}
%
A negative sample whose representation is close to the anchor sample provides a triplet that is more likely to produce non-zero loss. The main purpose of BoN is providing these relevant negative samples using an online algorithm that is loss-independent and computationally inexpensive. BoN is inspired by the Spectral Hashing method \cite{weiss2009spectral}. Nonetheless, we implemented several changes in order to efficiently adapt to the negative mining problem during training.

Spectral Hashing is a nearest neighbour search algorithm that is shown to be better performing than Product Quantization while being simpler to implement and more efficient at learning the hash function \cite{weiss2009spectral}. In terms of performance, it is inferior with respect to methods that address the embedding compression and the quantization as a whole problem, e.g. \cite{IterativeQuantization2011, BinaryAutoencoders2015}. However, we have to consider that the embedding is changing during the training, thus a simpler but flexible approach is preferred over methods providing better results at a greater computational cost.

The main approach of the Spectral Hashing method is to (1) learn a linear projection from a high dimensional space (of size $e$) to a lower dimensional space (of size $s \ll e$) by means of a standard PCA; (2) apply the projection to a sample; (3) perform a 1 bit quantization over every dimension by threshold at 0; and (4) group the $s$ bits into an integer codeword. The codeword represents the entry of an hash table. The underlying assumption is that samples falling into the same bin are neighbours in the high dimensional embedding. Of course, this assumption is overly optimistic and the variation from the optimal behavior is mainly due to the following facts: (1) since $s \ll e$ we lose some information about the topology of the high dimensional embedding and (2) the quantization is extremely harsh and there is no actual control over the quantization error during the process. Nonetheless, experimental validation shows that the Spectral Hashing method indeed performs well in retrieval tasks \cite{weiss2009spectral}.

However, the direct application of the Spectral Hashing (or any other nearest neighbour efficient algorithm) method to the problem of retrieving negative samples is not straightforward because the embedding is dynamically changing during the training. One can compute the whole embedding every certain number of steps, plus computing the PCA, and the hash table, but this na\"ive strategy does not scale well for large datasets. Consequently, we propose three main modifications to the Spectral Hashing approach in order to have an \textbf{online} algorithm that mimics its performance:
\begin{enumerate}
    \item The PCA is substituted by a linear auto-encoder paired with $L_2$ reconstruction loss.
    \item The quantization threshold is dynamically estimated, per dimension.
    \item The hash table is dynamically updated.    
\end{enumerate}

All three steps are computed over each input mini-batch at training time. Samples belonging to the mini-batch are updated in the hash table instantaneously. 
%
\subsection{Linear auto-encoder}
\label{sect:linear_AE}Since the online PCA estimation is in general computationally inefficient and potentially numerically unstable \cite{OnlinePCA2018}, we choose to train a linear autoencoder (AE) paired with $L_2$ reconstruction loss, as in formulas (\ref{eq:autoencoder}), where $h(x)$ is the projected subspace of dimensionality $s$. In order to not influence the embedding, we do not back-propagate the gradients generated by the $\mathcal{L}_{AE}$ loss through $f(x)$.
\begin{equation}
   \label{eq:autoencoder}
   \begin{aligned}
   h(x) = W_1 f(x) + b_1 \\
   \hat{f}(x) = W_2 h(x) + b_2 \\
   \mathcal{L}_{AE} = ||f(x) - \hat{f}(x) ||_2^2
   \end{aligned}
\end{equation} 

The AE continuously models the projection that provides the codeword to the hash table update procedure. The added cost of learning such an AE is negligible w.r.t. to the Siamese network training. We extract codewords of all samples in the mini-batch and we use them for hash table update in every step of training. At the same time we update the weights of the AE without backpropagating the gradients through the Siamese architecture. The only BoNs parameter is $s$, and its choice is related to two factors: (1) the smaller $s$, the more difficult to reconstruct (in the $L_2$ sense) the original sized embedding and (2) the bigger $s$, the larger the number of bins obtained after the binarization, more precisely $B=2^s$. A detailed analysis on the behaviour of BoN as a function of $s$ is presented in the experimental section.

\subsection{Dynamic quantization thresholds}
Since the lower dimensional space $h(x)$ changes dynamically during the training, the correct thresholds $\mu$ for its binarization is estimated as a running mean: $\mu \leftarrow \beta \mu + (1 - \beta) h(x)$ where $\beta$ controls how quickly the running average forgets old samples. We noticed that varying $\beta \in [0.95, 0.999]$ does not influence the results so that it is not a critical parameter to tune.
\subsection{Hash table dynamic update}
We maintain a hash table $L$ with $2^s$ bins that, for each entry indexed by an integer $j$, contains a collection of images identified by pairs $(v, ID(v))$, where $v$ is an integer that uniquely represents an image in the dataset, and $ID(v)$ is an integer that uniquely represents the ID for the given image. Also, we keep track of the latest hash entry for every image in the dataset, using integer values, such that $C[v] = j$. In such a way, updating the hash table has a very limited computational cost. The slowest part of the update procedure is removing the tuple $(v, ID(v))$ from the bin to which it had been previously assigned, and it has a cost of $\mathcal{O}(N / 2^s)$. In term of memory cost, assuming that both the ID and the sample identifiers ($ID(v)$ and $v$) can be represented with 4 bytes integers, we need only a total of $4 (N + 2N)$ bytes to store both the hash table $L$ and the list of hash entries $C$.

In each training step for each image in the mini-batch $I \in \{I_1, I_2, .., I_m\}$ we extract its codeword $c(I)$ by binarizing the difference between its latent projection from the autoencoder $h(I)$ and running mean $\mu$. We remove the pair $(v_I, ID(v_I))$ from the hash bin where it had been stored before $L[C[v_I]]$ and we add the pair to the new hash bin $L[c(I)]$. Finally, we update the list of hash entries $C[v_I]$ with the new hash entry $c(I)$.
\subsection{Bag of Negatives with random sampling}
The simplest way of using BoN is creating mini-batches of $b$ random anchor-positive image pairs. For each pair, we sample a negative image randomly among the images that belong to the same bin as the anchor. In case the anchor belongs to a bin in which there are no other images from a different ID, we sample the negative image randomly from the whole dataset.
\subsection{Bag of Negatives with \emph{batch hard} loss}
\label{sect:bonplusbatchhard}
As BoN can provide relevant images for batch sampling, it can be easily combined with a loss such as \textit{batch hard}. It is important to create batches of size $m$, which contain $k$ images from $l$ IDs when training a model with \textit{batch hard} loss. We set $k=2$ for all the experiments, as we are focusing on showing the importance of good negative sampling, and we want to avoid the results being influenced by hard positive samples. First we select a random non-empty bin, that contains images from $r$ different IDs. If $r$ is 1, we create a set of $l$ IDs randomly. If $r$ is greater or equal than $l$, we choose $l$ IDs randomly from the same bin. Finally, if $r$ is greater than 1 and smaller than $l$, we take all the $r$ IDs from the chosen bin, and pick another random bin from which we sample the remaining $l - r$ IDs, if possible. We keep repeating the last step until we have $l$ IDs. Once we have a set of $l$ IDs, we create a mini-batch of $k$ random images that belong to each one of $l$ selected IDs.
\section{Empirical evidence}
\label{sect:experiments}
\subsection{Datasets}
\textbf{Person re-identification large dataset.} We merged eleven publicly available datasets for person re-identification, CUHK01\cite{CUHK01}, CUHK02 \cite{CUHK02}, 3DPeS \cite{3DPeS}, VIPeR \cite{VIPeR}, airport\cite{resystematic}, MSMT17 \cite{MSMT}, Market-1501 \cite{Market1501}, DukeMTMC \cite{DukeMTMC}. The merged dataset has $178,002$ images of $10,552$ identities. We used both training and testing partitions of  all the datasets except for Market-1501 and DukeMTMC-reID and we did not use the images that are labeled as distructors or junk.

\textbf{Stanford Online Products} \cite{oh2016deep} is a retrieval dataset which contains $120,053$ images of $22,634$ products. The dataset is split into two partitions: the training partition, which contains $59,551$ images of $11,318$ products, and the testing partition which contains $60,502$ of $11,316$ classes.
\subsection{Pre-trained backbone and training parameters} 
\label{sect:settings} We use Inception-V3 as a backbone for our model. In particular, we take the convolutional layers and initialize them with weights from a standard network  pre-trained on ImageNet. The final descriptors are further max-pooled and $\ell_2$ normalized. The descriptor size is $2,048$. The model is trained using the ADAM optimizer, with the initial learning rate $10^{-4}$, and with a learning rate decay of $0.9$ each $50\textrm{k}$ iterations. The images for person re-ID are resized to $192 \times 384$ pixels, and $256 \times 256$ for Stanford Online Products. At test time, we extract representations and compare them using dot product.
\subsection{Analysis of Bag of Negatives}
\begin{figure}
\begin{tabular}{cc}
\bmvaHangBox{\parbox{0.45\linewidth}{\rule{0pt}{0ex}\vspace{1mm}

\begin{tikzpicture}[scale=0.7]
\begin{axis}[
    title={},
    xlabel={mAP on train set},
    ylabel={\% of non-zero loss triplets},
    xmin=27, xmax=100,
    ymin=0, ymax=80,
    xtick={0,10,...,100},
    ytick={0,10,...,70},
    style={thick},
    legend pos=north west,
    legend cell align={left},
    legend style={fill=none},
    ymajorgrids=true,
    xmajorgrids=true,
    grid style=dashed,
]
 
\addplot[
    color=blue,
    ]
    coordinates {
    (28.7163373252,18.4)(36.6703044014,14.4)(40.5213232793,13.2)(44.1490478501,11.6)(46.1420791893,11.1)(45.8713661335,9.5)(47.6258896295,9.1)(50.152689209,9.7)(50.6922203085,8.2)(51.6554580368,7.9)(53.4388685123,7.1)(55.0778191997,6.5)(56.4574105954,6.7)(56.9870754057,6.7)(57.848971106,6.9)(59.0049069331,5.9)(59.67555673,6.7)(60.3804160771,5.5)(61.3693872119,5.6)(63.0079798889,5.3)(63.1897775142,5.9)(64.1196821525,4.6)(63.9346406276,5.1)(64.9467644072,5.5)(65.3038896583,4.6)(65.2627150295,5.2)(66.3579656012,4.6)(66.7779509224,5.2)(66.9773998035,5.1)(67.8055931258,4.4)(67.7426224877,4.1)(69.2329814353,4.5)(69.963570855,5.0)(70.3120673133,4.0)(70.3046882905,4.2)(70.4342710494,4.5)(70.4229726818,4.6)(71.0352347046,4.5)(71.8294601902,3.7)(72.7955501815,3.8)(73.1910261495,4.0)(73.444905857,4.2)(73.7142466027,3.5)(74.2353449174,3.8)(74.0550374064,4.0)(73.8845618503,3.2)(74.5434961856,3.5)(74.8983744928,3.4)(75.4157902608,3.4)(75.8426757714,3.5)(75.7089458136,3.3)(75.6142812929,3.6)(75.9231932426,3.0)(76.172261676,3.2)(76.5103638125,3.2)(76.6136700111,3.0)(76.9178202751,3.6)(77.5902465068,3.3)(77.4338733006,3.3)
    };
    
    
\addplot[
    color=red,
    ]
    coordinates {
    (44.0306164484,50.7)(57.3235417207,47.7)(65.4880947669,45.9)(71.0567404669,42.8)(75.5560288646,39.7)(77.3463403132,38.9)(79.175825302,37.9)(80.4534890259,36.6)(83.0121777344,35.2)(84.3149309332,35.8)(85.6581852596,32.7)(86.822729957,31.9)(87.3020640417,32.8)(88.0838345516,31.3)(88.9113429832,34.0)(89.681764169,30.0)(90.2541316048,31.9)(90.4381562129,31.5)(90.7518244809,27.5)(90.9841910558,28.9)(91.6034213005,29.4)(92.3652478043,29.5)(92.7680955196,27.2)(92.9991555733,26.8)(93.3811144393,25.5)(93.5495105492,27.3)(93.5441839693,27.3)(93.4834192592,27.9)(93.7082245414,25.6)(94.2002869127,27.5)(94.3429513491,25.1)(94.5657898919,25.2)(94.5612958961,26.2)(94.6689689517,24.2)(94.8792309695,24.8)(94.8584244814,23.2)(95.0913953278,23.7)(95.1778334702,23.6)(95.2311398793,23.2)(95.3048457246,23.7)(95.5081645747,23.0)(95.5769352778,23.4)(95.5920313264,21.6)(95.7565969938,23.3)(95.8621249026,26.8)
    };
    
\addplot[
    color=green,
    ]
    coordinates {
    (63.703749652,51.0)(75.5085732544,40.7)(79.4265571756,33.9)(83.5569304704,29.8)(85.6113900647,26.8)(87.635781043,23.7)(89.2812988677,21.4)(89.7691645066,20.2)(90.6103739445,22.6)(91.1865631677,19.8)(91.817997293,17.6)(92.3333808179,19.2)(92.8267528187,16.0)(93.0975907104,16.9)(93.1328749377,15.7)(93.5339432715,15.3)(93.8272545208,14.4)(94.2146070054,14.3)(94.5602811738,14.7)(94.6297419676,13.1)(94.8975947105,11.7)(95.0743731202,12.5)(95.2541104722,11.4)(95.364736897,11.3)(95.2602447686,10.6)(95.9582844538,10.4)(96.2543796359,11.3)(96.4189837969,11.4)(96.4846762115,10.6)(96.4780743617,11.5)(96.712333015,10.0)(96.4449986207,10.4)(96.7864100581,8.9)(96.620082213,9.4)(96.7012790498,9.7)(96.7712950551,13.7)
    };
    
\addplot[
    color=orange,
    ]
    coordinates {
    (83.6659668576,37.7)(89.0454972832,26.6)(92.4220082929,21.4)(93.8037572704,19.5)(95.3955654571,17.2)(95.8834720443,14.2)(96.8036538454,13.6)(97.0801500895,11.9)(97.4642116208,11.1)(97.6247233469,10.3)(98.0490964236,9.6)(98.537461471,8.7)(98.6442501108,8.3)(98.6918904425,9.3)(98.768299211,7.6)(98.7819139185,6.5)(98.9698809032,6.0)(99.1946136883,5.9)(99.3893469478,6.2)(99.1866626135,6.7)(99.3295151365,5.6)(99.2354035573,5.6)(99.3655560021,4.9)(99.3722313175,4.4)(99.2760294005,4.2)(99.3305624007,5.0)(99.3622859484,4.3)(99.54448397,4.1)(99.359366498,3.8)(99.3304584729,3.6)(99.3472443544,3.5)
    };
    
\addplot[
    color=black,
    ]
    coordinates {
    (81.695709631,70.7)(87.7022528807,59.2)(90.7363416898,49.7)(92.8442284219,43.9)(94.5166976224,38.2)(95.4371191987,33.4)(96.2807929852,27.0)(96.7877702484,25.8)(97.9693517018,23.4)(98.2934889892,20.5)(99.1100129806,18.0)(99.3349897299,17.6)(99.4914042216,15.7)(99.5240176549,14.9)(99.4585012832,14.4)(99.4349164766,12.0)(99.5095069289,10.6)(99.3936855984,10.0)(99.5374370926,9.9)(99.5982723109,8.8)(99.7676682775,8.5)(99.6297833081,8.5)(99.501029873,7.8)(99.762688746,7.6)(99.9593791779,6.5)(99.9571133469,6.5)(99.8100517335,5.6)(99.8095001364,5.7)(99.8276382921,5.5)(99.8031472565,5.0)(99.8179023018,4.5)(99.9775981445,4.4)(99.9627214133,4.8)(99.9861129993,4.6)(99.9685730563,3.3)(99.9566084146,3.6)(99.8311992607,3.5)(99.9645597248,3.6)(99.8558773737,3.9)(99.9951326124,3.3)(99.9861543763,2.9)(99.983317309,2.6)(99.992635447,3.0)(99.9912180951,3.2)(99.9879585763,2.7)(99.9965834548,2.8)(99.9786857012,2.5)(99.9961877338,1.9)(99.992239726,2.4)(99.960752958,2.6)(99.9850686366,2.2)(99.9885802857,2.0)(99.9936783305,1.8)(99.9677033843,1.7)(99.9732142857,2.0)(99.9732142857,1.5)(99.9792729592,1.7)(100.0,1.8)(100.0,1.8)(100.0,1.9)(100.0,1.4)(100.0,1.6)(100.0,2.1)(100.0,1.4)(99.9370596453,1.4)(99.9331116375,1.7)(99.9613245901,1.3)(100.0,1.2)(99.9826388889,1.1)(100.0,1.5)(100.0,1.2)(99.9965834548,0.8)(99.9965834548,1.2)(100.0,0.7)(100.0,1.0)
    };
    
\addplot[
    color=magenta,
    ]
    coordinates {
    (79.695472075,79.4)(87.7787806068,69.9)(92.5389643534,61.3)(94.4233837367,56.5)(96.2794670895,46.3)(97.3665467802,37.9)(98.1915863357,34.7)(98.4889402085,33.8)(98.9378541823,29.5)(99.3438862593,26.7)(99.6005413231,19.5)(99.7001463097,22.0)(99.7714944499,23.9)
    };
\addplot[
    color=gray,
    ]
    coordinates {
    (68.8515798922,55.2)(81.4455623295,46.7)(87.2213730151,38.8)(90.5637578769,34.3)(93.4108463406,28.2)(94.5457354766,24.0)(95.945822043,22.2)(96.4794941024,21.3)(97.1347093292,17.6)(97.7290236391,17.3)(98.1116590221,15.1)(98.3238279636,11.2)};
    \legend{Vanilla, BoN, \textit{semi hard}, \textit{batch hard}, BoN + BH, SH + BH, 100k IDs}
\end{axis}
\end{tikzpicture}}}&
\bmvaHangBox{\parbox{0.45\linewidth}{\rule{0pt}{0ex}\begin{tikzpicture}[scale=0.7] 
\begin{axis}[
    xlabel={number of steps[$\times 1000$]},
    ylabel={mAP[\%]},
    xmin=0, xmax=300,
    ymin=0, ymax=100,
    xtick={0,50,...,300},
    ytick={0,20,40,60,80,100},
    style={thick},
    legend pos=south east,
    legend cell align={left},
    legend style={fill=none},
    ymajorgrids=true,
    xmajorgrids=true,
    grid style=dashed,
]
 
\addplot[
    color=blue,
    ]
    coordinates {
    (10,28.7)(20,36.7)(30,40.5)(40,44.1)(50,46.1)(60,45.9)(70,47.6)(80,50.2)(90,50.7)(100,51.7)(110,53.4)(120,55.1)(130,56.5)(140,57.0)(150,57.8)(160,59.0)(170,59.7)(180,60.4)(190,61.4)(200,63.0)(210,63.2)(220,64.1)(230,63.9)(240,64.9)(250,65.3)(260,65.3)(270,66.4)(280,66.8)(290,67.0)(300,67.8)
    };
    
\addplot[
    color=red,
    ]
    coordinates {
    (10,44.0)(20,57.3)(30,65.5)(40,71.1)(50,75.6)(60,77.3)(70,79.2)(80,80.5)(90,83.0)(100,84.3)(110,85.7)(120,86.8)(130,87.3)(140,88.1)(150,88.9)(160,89.7)(170,90.3)(180,90.4)(190,90.8)(200,91.0)(210,91.6)(220,92.4)(230,92.8)(240,93.0)(250,93.4)(260,93.5)(270,93.5)(280,93.5)(290,93.7)(300,94.2)(310,94.3)(320,94.6)(330,94.6)(340,94.7)(350,94.9)(360,94.9)(370,95.1)(380,95.2)(390,95.2)(400,95.3)(410,95.5)(420,95.6)(430,95.6)(440,95.8)(450,95.9)
    };
    
\addplot[
    color=green,
    ]
    coordinates {
    (10,63.7)(20,75.5)(30,79.4)(40,83.6)(50,85.6)(60,87.6)(70,89.3)(80,89.8)(90,90.6)(100,91.2)(110,91.8)(120,92.3)(130,92.8)(140,93.1)(150,93.1)(160,93.5)(170,93.8)(180,94.2)(190,94.6)(200,94.6)(210,94.9)(220,95.1)(230,95.3)(240,95.4)(250,95.3)(260,96.0)(270,96.3)(280,96.4)(290,96.5)(300,96.5)
    };
    
\addplot[
    color=orange,
    ]
    coordinates {
    (10,83.7)(20,89.0)(30,92.4)(40,93.8)(50,95.4)(60,95.9)(70,96.8)(80,97.1)(90,97.5)(100,97.6)(110,98.0)(120,98.5)(130,98.6)(140,98.7)(150,98.8)(160,98.8)(170,99.0)(180,99.2)(190,99.4)(200,99.2)(210,99.3)(220,99.2)(230,99.4)(240,99.4)(250,99.3)(260,99.3)(270,99.4)(280,99.5)(290,99.4)(300,99.3)
    };
    
\addplot[
    color=black,
    ]
    coordinates {
    (10,83.8)(20,89.4)(30,91.5)(40,93.6)(50,94.8)(60,95.2)(70,96.0)(80,96.5)(90,97.0)(100,97.2)(110,96.7)(120,97.8)(130,98.1)(140,98.7)(150,98.5)(160,98.9)(170,98.8)(180,99.2)(190,99.4)(200,99.4)(210,99.7)(220,99.6)(230,99.6)(240,99.6)(250,99.7)(260,99.7)(270,99.7)(280,99.9)(290,99.7)(300,99.9)(310,99.9)(320,99.8)(330,99.7)(340,99.8)(350,99.8)(360,99.8)(370,99.8)(380,99.9)(390,99.9)(400,99.9)(410,99.9)(420,99.8)(430,99.8)(440,99.9)(450,99.9)(460,100.0)(470,100.0)
    };
\addplot[
    color=magenta,
    ]
    coordinates {
    (10,79.7)(20,87.8)(30,92.5)(40,94.4)(50,96.3)(60,97.4)(70,98.2)(80,98.5)(90,98.9)(100,99.3)(110,99.6)(120,99.7)(130,99.8)(140,99.8)(150,99.8)(160,99.8)(170,99.8)(180,99.9)(190,99.9)(200,99.9)(210,100.0)(220,100.0)(230,100.0)(240,100.0)(250,100.0)(260,100.0)(270,100.0)(280,100.0)(290,100.0)(300,100.0) 
    };
\addplot[
    color=gray,
    ]
    coordinates {
   (10,68.9)(20,81.4)(30,87.2)(40,90.6)(50,93.4)(60,94.5)(70,95.9)(80,96.5)(90,97.1)(100,97.7)(110,98.1)(120,98.3)(300,98.3)};

\addplot[dotted,
    color=blue,
    ]
    coordinates {
    (10,6.0)(20,7.4)(30,8.6)(40,9.2)(50,10.0)(60,10.4)(70,11.4)(80,12.4)(90,13.0)(100,13.7)(110,14.2)(120,14.8)(130,15.9)(140,16.3)(150,16.7)(160,17.3)(170,17.9)(180,18.5)(190,18.9)(200,18.6)(210,19.1)(220,19.8)(230,20.3)(240,20.6)(250,21.7)(260,21.4)(270,21.6)(280,22.3)(290,22.3)(300,22.5)
    };
    
\addplot[dotted,
    color=red,
    ]
    coordinates {
    (10,11.3)(20,19.4)(30,26.2)(40,31.1)(50,35.1)(60,38.0)(70,40.4)(80,42.8)(90,45.1)(100,46.3)(110,47.8)(120,48.9)(130,50.0)(140,50.7)(150,51.7)(160,52.4)(170,53.3)(180,54.1)(190,54.8)(200,55.3)(210,55.9)(220,56.7)(230,57.0)(240,57.2)(250,58.2)(260,58.4)(270,58.5)(280,59.0)(290,59.4)(300,59.5)(310,59.5)(320,59.9)(330,59.9)(340,60.2)(350,60.3)(360,60.7)(370,60.5)(380,60.5)(390,60.4)(400,60.7)(410,60.7)(420,61.2)(430,61.1)(440,61.4)
    };
    
\addplot[dotted,
    color=green,
    ]
    coordinates {
    (10,25.8)(20,34.1)(30,39.4)(40,42.5)(50,44.8)(60,46.9)(70,48.6)(80,49.8)(90,50.7)(100,52.0)(110,52.6)(120,53.0)(130,53.3)(140,53.9)(150,54.1)(160,54.7)(170,55.5)(180,55.5)(190,55.3)(200,55.8)(210,56.5)(22
0,57.0)(230,57.1)(240,57.1)(250,57.2)(260,57.0)(270,57.0)(280,57.8)(290,58.0)(300,58.0)
    };
    
\addplot[dotted,
    color=orange,
    ]
    coordinates {
    (10,40.4)(20,47.7)(30,51.8)(40,53.2)(50,54.7)(60,56.1)(70,56.3)(80,56.9)(90,57.8)(100,58.2)(110,58.5)(120,58.6)(130,58.7)(140,58.7)(150,59.0)(160,59.1)(170,59.3)(180,59.7)(190,60.0)(200,60.0)(210,60.1)(22
0,60.0)(230,60.2)(240,60.3)(250,60.6)(260,60.6)(270,60.8)(280,60.7)(290,60.6)(300,60.9)
    };
    
\addplot[dotted,
    color=black,
    ]
    coordinates {
    (10,49.0)(20,60.9)(30,65.0)(40,67.0)(50,68.4)(60,68.9)(70,69.1)(80,69.5)(90,69.5)(100,69.2)(110,69.2)(120,69.2)(130,69.5)(140,69.1)(150,69.2)(160,69.1)(170,69.2)(180,68.8)(190,68.6)(200,68.8)(210,68.7)(220,68.7)(230,68.5)(240,68.5)(250,68.1)(260,67.8)(270,67.5)(280,67.7)(290,67.6)(300,67.7)(310,67.7)(320,67.6)(330,67.6)(340,67.6)(350,67.6)(360,67.6)(370,67.7)(380,67.4)(390,67.3)(400,67.3)(410,67.4)(420,67.4)(430,67.3)(440,67.2)(450,67.1)(460,67.1)(470,67.0)
    };
    
\addplot[dotted,
    color=magenta,
    ]
    coordinates {
    (10,48.4)(20,63.3)(30,67.5)(40,70.1)(50,71.2)(60,71.3)(70,71.5)(80,71.6)(90,71.6)(100,71.7)(110,71.5)(120,71.2)(130,71.0)(140,70.8)(150,70.2)(160,70.3)(170,70.2)(180,70.5)(190,70.1)(200,69.7)(210,69.8)(220,69.9)(230,69.7)(240,69.7)(250,69.6)(260,69.7)(270,69.4)(280,69.7)(290,69.7)(300,69.5) 
    };
    
\addplot[dotted,
    color=gray,
    ]
    coordinates {
    (10,44.9)(20,58.4)(30,62.9)(40,64.9)(50,66.0)(60,66.8)(70,67.3)(80,67.7)(90,67.8)(100,67.6)(110,67.4)(120,67.4)(300, 67.2)
    };
 
    \legend{Vanilla, BoN, \textit{semi hard}, \textit{batch hard}, BoN + BH, SH + BH, 100k IDs}

\end{axis}
\end{tikzpicture}}}\\
(a)&(b)
\end{tabular}
\caption{a)  Percentage of non-zero loss triplets as a function of mAP on the training set. b) Models evaluated on Training and Market datasets as functions of number of steps.}
\label{fig:map_plots}
\end{figure}
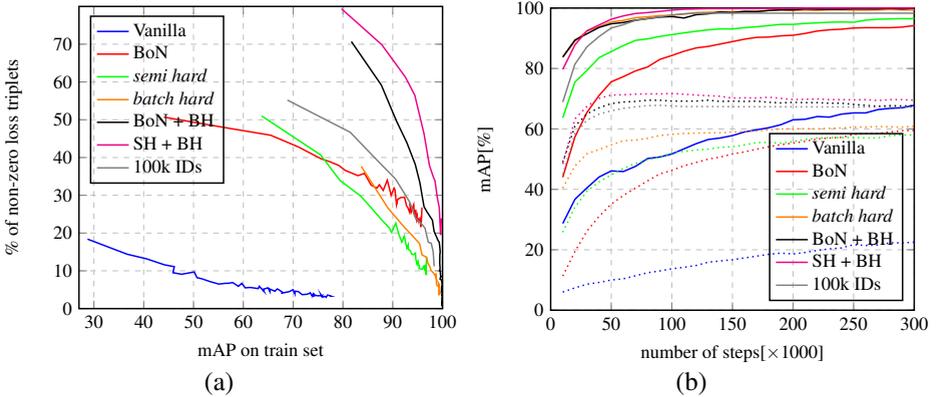
\subsubsection{Non-zero loss triplets analysis}
Figure \ref{fig:map_plots}.a shows the percentage of non-zero loss triplets (measured at training time) as a function of the training mean Average Precision (mAP) for the Vanilla sampling, BoN-Random, the \emph{semi hard} loss, the \textit{batch hard} loss, BoN-\textit{batch hard}, Spectral Hashing-\textit{batch hard} and 100k IDs-\textit{batch hard} loss on the person re-ID dataset. The margin is set to $\alpha = 0.3$, the mini-batch size is $m=48$, and the leftmost point on the plot for every method is obtained at 10k steps of training in all experiments.
As expected, the percentage of non-zero triplets for the Vanilla sampling starts at only $20\%$ and decreases as the mAP increases; at $\text{mAP}=77.4$ the non-zero triplets are less than $5\%$ and the training is virtually unable to learn anything else. 

BoN-random significantly increases the number of non-zero loss triplets w.r.t. the Vanilla sampling, without modifying the loss nor the way the anchor-positive pairs are formed. The improvement is solely due to the improved sampling of negatives. BoN-random exhibits a behaviour similar to \textit{semi hard} and \textit{batch hard} while providing, in general, more non-zero loss triplets. However, the nature of the improvement provided by our method and \textit{batch hard} is very different: BoN searches for negatives in a local region of the embedding space while \textit{batch hard} forms the triplets seeking non-zero loss triplets in an explicit way within a mini-batch but sampling from the whole embedding.

These two complementary strategies can be easily combined as seen in section \ref{sect:bonplusbatchhard}. The combination inherits the benefits of both approaches: at 10k steps \emph{batch hard} and BoN-\emph{batch hard} have a similar $\textrm{mAP} \approx 83\%$ but BoN-\emph{batch hard} has about 2 times more non-zero loss triplets, and it has more non-zero loss triplets systematically until the end of the training. As will be seen in the comparison, this behavior not only speeds-up the training, but also provides better triplets, which improves the performance on validation sets.

The combination of Spectral Hashing and \textit{batch hard} requires the following steps: (1) feature extraction on the whole training set, (2) reduction of the feature size by PCA to the size $s$ ($s = 18$) and (3) hash table construction; we repeat this procedure every 5k steps. Given this hash table, batches are created the same way as explained in \ref{sect:bonplusbatchhard}. BoN-\textit{batch hard} shows very similar behavior to the Spectral Hashing - \textit{batch hard}: they both train quickly, obtaining almost the same mAP after 10k steps, with high percentage of non-zero loss triplets. Spectral Hashing - \textit{batch hard} is providing more non-zero loss triplets during the whole training. This is expected, as the hash table is updated at the same moment for all the samples. However, this configuration is far more computationally expensive and does not scale for datasets with a large number of IDs. As an example, each hash table update requires $11.6$ minutes of computation every 5k steps, which is the same time required for $1,657$ steps of training. In other words, $25\%$ of the training time is spent on hash table update. Thus SH-BH is better, but at a much higher computational cost.

We analyze the behavior of batch creation proposed in \cite{100Kids2017}, using 10 clusters as suggested by the authors. We use these clusters for creating the hash table and we do not update it during the training. In addition to longer training time, this method lacks flexibility in updating the hash table. In other words, samples that are considered relevant negatives to an identity are set at the beginning of the training and are static w.r.t. the training process. Moreover, a possible sub-optimal clustering is going to be seriously detrimental to the training.
\subsubsection{BoN-Random behaviour varying $s$}
\begin{figure}
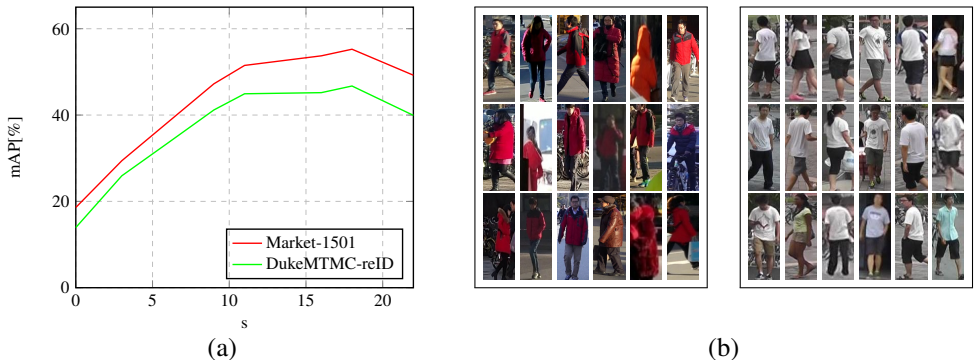

\begin{tabular}{ccc}
\bmvaHangBox{\parbox{0.45\linewidth}{\rule{0pt}{0ex}\begin{tikzpicture}[scale=0.65]
\begin{axis}[
    title={},
    xlabel={s},
    ylabel={mAP[\%]},
    xmin=0, xmax=22,
    ymin=0, ymax=65,
    xtick={0,5,...,22},
    ytick={0,20,40,60,80, 100},
    style={thick},
    legend pos=south east,
    legend cell align={left},
    legend style={fill=none},
    ymajorgrids=true,
    xmajorgrids=true,
    grid style=dashed,
]
    
\addplot[
    color=red,
    ]
    coordinates {(0,18.59)(3,29.42)(6,38.33)(9,47.24)(11,51.5)(16,53.72)(18,55.25)(22,49.23)
    
    };
    
\addplot[
    color=green,
    ]
    coordinates {(0,13.94)(3,25.93)(6,33.53)(9,41.18)(11,44.93)(16,45.19)(18,46.73)(22,39.93)
    
    };

    \legend{Market-1501, DukeMTMC-reID}
 
\end{axis}
\end{tikzpicture}}}&
\bmvaHangBox{\fbox{\parbox{0.22\linewidth}{\rule{0pt}{0ex}\input{bin5.tex}}}}&
\bmvaHangBox{\fbox{\parbox{0.22\linewidth}{\rule{0pt}{0ex}\input{bin2.tex}}}}\\
(a)&\multicolumn{2}{c}{(b)}
\end{tabular}
\caption{a) Validation mAP as a function of $s$, b) Samples from 2 random bins.}
\label{fig:dist}
\end{figure}

 It is interesting to note that BoN-Random degenerates to Vanilla sampling for $s=0$. Figure \ref{fig:dist}.a shows the mAP results on the Market and Duke validation datasets for different values of $s$ at 200k steps. As it can be seen, the performance increases with $s$ and it reaches a maximum at $s = 18$; with this value we have $178k / 2^{18} = 0.68$ samples per bin on average. Within a static embedding, and with the standard Spectral Hashing, these figures represent degenerate behaviour in which every bin contains at most one sample, thus defeating the objective of the hashing in providing negative candidates. In reality, the embedding, the linear projection and the thresholds are changing dynamically, such that many bins can be empty. In this case, increasing $s$ improves the granularity of the hash table without incurring in degenerate hash functions. Nonetheless, with $s=22$, BoN reaches its breaking point and the average number of samples per bins (for non empty bins) is very high, such that BoN-Random starts to perform negative sampling on the whole dataset too frequently.

\subsubsection{Qualitative results}
In Figure \ref{fig:dist}.b we show images sampled from 2 bins for the person re-identification dataset for BoN-random at 200k steps. As demonstrated, the images belonging to the same bin have some similarity, e.g. same upper body color, same type of clothes etc.
\section{Results and Comparison}
In this section we perform a controlled comparison of our proposal with the two most commonly used triplet losses: \emph{semi hard} and \emph{batch hard} and the two batch creation methods: hierarchical tree \cite{ge2018deep} and 100k IDs \cite{100Kids2017}. We avoid extra variables (e.g. augmentation, other architectures, etc.) that could mask the empirical results for other reasons not related to negative sampling and triplet construction. For such reasons, we use the same mini-batch size for all the methods, the same pre-trained backbone, the same margin $\alpha$ and the same embedding size (see subsection \ref{sect:settings} for the details).

Table \ref{tab:results} shows the results of the comparison on the person re-identification dataset. BoN-random clearly outperforms Vanilla sampling and provides validation mAPs comparable to \emph{semi hard} and \emph{batch hard}. The best results are obtained by Spectral Hashing - \emph{batch hard}, which was expected, taking into account that BoN is an online approximation of Spectral Hashing. The numbers show that the margin between BoN and Spectral Hashing is only 1.5\% on average of the two evaluation datasets. However, Spectral Hashing can be used only if the training set is small, so its application on bigger datasets would be infeasible. One can argue that the performance of BoN can be easily reached by just increasing the mini batch size of the \emph{batch hard} method. The "\textit{batch hard} (2x batch)" line shows results from the training in which the mini-batch size is doubled. As expected, in this case, the model trains faster (in terms of number of steps until convergence) and has better performance, but does not outperform BoN-\emph{batch hard}. This experiment shows that BoN is a key component of the accelerated training and improved validation results of BoN-\emph{batch hard}.

We implemented two methods for batch creation known in the literature, Hierarchical Tree (HT) \cite{ge2018deep} and 100k IDs \cite{100Kids2017} and combined them with \textit{batch hard}. We followed the procedure described in \cite{ge2018deep} and computed the distance matrix between all the IDs every 5k steps. We form a batch by randomly selecting one ID, and taking the remaining $l - 1$ as its closest neighbors. Even though Hierarchical Tree and 100k IDs require a computationally expensive two-phase training, BoN-\textit{batch hard} obtains better results in fewer training steps.

We fine-tuned the BoN-\emph{batch hard} model from the table \ref{tab:results} on Market-1501 and DukeMTMC-reID datasets. To the best of our knowledge, the fine tuned BoN-\textit{batch hard} performs better than all the methods that do not use ad-hoc architectures.

Figure \ref{fig:map_plots}.b shows the training (full lines) and validation (dashed lines) performance of the compared methods up to 300k steps. BoN-random exhibits the slower increasing curve in all plots with respect to all methods except the Vanilla sampling. This is due to the fact that the negative is still sampled randomly inside the anchor bin. Nonetheless, it is as good as \textit{semi hard} and \textit{batch hard} on the validation sets at 300k, which means that the negative samples are of sufficiently good quality to provide good generalization capability. BoN-\emph{batch hard} is the fastest method, reaching the over-fitting regime at about 70k-100k steps; this is at least 3 times faster than other methods.

Table \ref{tab:stanford} shows the comparison of BoN-\emph{batch hard} with state-of-the-art approaches on Stanford Online Products dataset. We trained BoN-\emph{batch hard} using the same methodology as explained in section \ref{sect:settings} on retrieval, using data augmentation techniques such as random horizontal flipping, blurring, zooming in and out, and cutout. BoN-\emph{batch hard} obtains the state-of-the-art results when compared with all the other methods.

\emph{Batch hard} and BoN are complementary: BoN provides a set of good quality negative samples, while \emph{batch hard} provides the explicit hard negative selection process and the increased number of triplets per mini-batch. The resulting combination is faster and better than both approaches.
\begin{table}[t]
\begin{minipage}[b]{.55\linewidth}
\caption{mAP validation results at peak performance for every method. * stands for the best state-of-the-art result obtained by an ad-hoc architecture.}
\label{tab:results}
\begin{center}
\begin{small}
\begin{tabular}{cccc}
\toprule
Method            & \#steps & Market & Duke \\
\midrule
Vanilla           & 600k    & 28.1   & 22.5 \\
BoN-random        & 440k    & 61.4   & 51.3 \\
\emph{semi hard}  & 240k    & 57.0   & 50.5 \\
\emph{batch hard} & 280k    & 60.8   & 53.7 \\
HT~\cite{ge2018deep}-\emph{batch hard} & 310k    &  65.9  & 57.5 \\
100k~\cite{100Kids2017}-\emph{batch hard} & 90k    &  67.8  & 61.2 \\
BoN-\emph{batch hard} & \textbf{80k} & \textbf{69.5} & \textbf{62.1} \\
\midrule
SH-\emph{batch hard} & 100k & 71.7 & 62.9 \\
\emph{batch hard} (2x batch) & 70k & 62.92 & 56.76 \\
\midrule
SPReID \cite{Kalayeh_2018_CVPR} & - & 76.6 & 63.3 \\
MGN \cite{wang2018learning}* & - & 86.9 & 78.4 \\
BoN-\emph{batch hard}- ft & - & 77.3 & 68.6 \\
\bottomrule
\end{tabular}
\end{small}
\end{center}
\end{minipage}
\begin{minipage}[b]{.4\linewidth}
\caption{Results on Stanford Online Products \cite{oh2016deep}. * stands for the best state-of-the-art result obtained by an ad-hoc architecture.}
\label{tab:stanford}
\begin{center}
\begin{small}
\begin{tabular}{cccc}
\toprule
Method                  & r1 & r10 \\
\midrule
lifted structured \cite{oh2016deep}        & 61.5 & 80.0 \\
sampling matters \cite{samplingmatters2017iccv}  & 72.7 & 86.2 \\
hierarchical tree \cite{ge2018deep}          & 74.8 & 88.3 \\
ABE-$8^{512}$ \cite{kim2018attention}*        & 76.3 & 86.4 \\
\midrule
BoN-\emph{batch hard} & \textbf{75.8} & \textbf{88.6} \\
\bottomrule
\end{tabular}
\end{small}
\end{center}
\end{minipage}
\vspace{-0.5cm}
\end{table}
\section{Conclusion}
\vspace{-0.25cm}
In this paper we introduced Bag of Negatives (BoN), a novel method for accelerated  and  improved  training  of  Siamese networks that scales linearly on datasets with large numbers of identities. The method is complementary to the popular \emph{batch hard} approach, and their combination provides improved validation results by means of high quality negative candidates. Future work will address a profound analysis of BoN behaviour when combined with other loss functions. Also, a more extensive test on other datasets will be performed. Finally, we will investigate possible solutions on automated estimation of the $s$ meta-parameter.

\section*{Acknowledgements}
\vspace{-0.25cm}
This work has been partially supported by the Spanish project RTI2018-095645-B-C21, the grant 2018FI\_B1\_00056 from AGAUR and CERCA, from Generalitat de Catalunya.

\bibliography{biblio}

\end{document}